\documentstyle[10pt,colap]{article}
\author{Flora Ram\'{\i}rez Bustamante \\
Escuela Polit\'{e}cnica Superior \\
Universidad Carlos III de Madrid \\
c/ Butarque 15 \\
28011 Legan\'{e}s, Madrid, SPAIN \\
{\tt flora@ia.uc3m.es} \And
Fernando S\'{a}nchez Le\'{o}n \\
Laboratorio de Ling\"u\'{\i}stica Inform\'atica \\
Facultad de Filosof\'{\i}a y Letras \\
Universidad Aut\'onoma de Madrid \\
28049 Madrid, SPAIN \\
{\tt fernando@maria.lllf.uam.es}}
\title{GramCheck: A Grammar and Style Checker}

\begin{document}

\maketitle

\begin{abstract}
This paper presents a grammar and style checker demonstrator
for Spanish and Greek native writers developed within the project
GramCheck. Besides a brief grammar error typology for Spanish, a
linguistically motivated approach to detection and diagnosis is
presented, based on the generalized use of PROLOG extensions to highly
typed unification-based grammars. The demonstrator, currently including
full coverage for agreement errors and certain head-argument relation
issues, also provides correction by means of an
analysis-transfer-synthesis cycle. Finally, future extensions to the
current system are discussed.
\end{abstract}

\section{Introduction}

Grammar checking stemmed as a logical application from former attempts
to natural language understanding by computers. Many of the NLU systems
developed in the 70's included a kind of error recovery mechanism
ranging from the treatment only of spelling errors,
PARRY~\cite{Parkinson77}, to the inclusion also of incomplete input
containing some kind of ellipsis, LADDER/LIFER~\cite{Hendrix77}.

The interest in the 80's begun to turn considering grammar checking as
an enterprise of its own right \cite{Carbonell83}, \cite{Hayes81},
\cite{Heidorn82}, \cite{Jensen83}, though many of the approaches were
still in the NLU tradition \cite{Charniak83}, \cite{Granger83},
\cite{Kwasny81}, \cite{Weischedel80}, \cite{Weischedel83}. A 1985 Ovum
report on natural language applications \cite{Johnson85} already
identifies grammar and style checking as one of the seven major
applications of NLP.  Currently, every project in grammar checking has
as its goal the creation of a writing aid rather than a robust
man-machine interface \cite{Adriaens94}, \cite{Bolioli92},
\cite{Vosse92}.

Current systems dealing with grammatical deviance have been mainly
involved in the integration of special techniques to detect and correct,
when possible, these deviances. In some cases, these have been
incorporated to traditional parsing techniques, as it is the case with
feature relaxation in the context of unification-based formalisms
\cite{Bolioli92}, or the addition of a set of catching error rules
specially handling the deviant constructions \cite{Thurmair90}. In
other cases, the relaxation component has been included as a new add-in
feature to the parsing algorithm, as in the IBM's PLNLP approach
\cite{Heidorn82}, or in the work developed for the Translator's
Workbench project using the METAL MT-system \cite{TWB92}.

Besides, an increasing concern in current projects is that of linguistic
relevance of the analysis performed by the grammar correction system. In
this sense, the adequate integration of error detection and correction
techniques within mainstream grammar formalisms has been addressed by a
number of these projects \cite{Bolioli92}, \cite{Vosse92},
\cite{Genthial92}, \cite{Genthial94}. \\

Following this concern, this paper presents results from the project
GramCheck (A Grammar and Style Checker, MLAP93-11), funded by the CEC.
GramCheck has developed a grammar checker demonstrator for Spanish and
Greek native writers using ALEP~\cite{ET6/1},~\cite{Simpkins94} as the
NLP development platform, a client-server architecture as implemented
in the X Windows system, Motif as the `look and feel' interface and
Xminfo as the knowledge base storage format. Generalized use of
extensions to the highly typed and unification based formalism
implemented in ALEP has been performed.  These extensions (called
Constraint Solvers, CSs) are nothing but pieces of PROLOG code
performing different boolean and relational operations over feature
values. Besides, GramCheck has used ongoing results from LS-GRAM
(LRE61029), a project aiming at the implementation of middle coverage
ALEP grammars for a number of European languages.

The demonstrator checks whether a document contains grammar errors or
style weaknesses and, if found any, users are provided with messages,
suggestions and, for grammar errors only, automatic correction(s).

\section{Brief grammar error typology for Spanish}

The linguistic statements made by developers of current grammar
checkers based on NLP techniques are often contradictory regarding the
types of errors that grammar checkers must correct automatically.
\cite{Veronis88} claims that native writers are unlikely to produce
errors involving morphological features, while \cite{Vosse92} accepts
such morpho-syntactic errors, in spite of the fact that an examination
of texts by the author revealed that their appearance in native
writer's texts is not frequent.  Both authors agree in characterizing
morpho-syntactic errors as a sample of lack of competence.

On the other hand, an examination of real texts produced by Spanish
writers revealed that they do produce morpho-syntactic
errors\footnote{The corpus used contains nearly 70,000 words including
text fragments from literature, newspapers, technical and administrative
documentation. It has been provided to a large extent by GramCheck pilot
user, ANAYA, S.A.}. Spanish is an inflectional language, which increases
the possibilities of such errors.  Nevertheless, other errors related to
structural configuration of the language are produced as well.

Errors found fall into one of the following subtypes, assuming that
featurization is the technique used in parsing sentences:

\begin{enumerate}

\item
Mismatching of features that do not affect representational issues
(intra- or inter-syntactic agreement on gender, number, person and case
for categories showing this phenomenon).  These mismatchings produce
{\em non-structural errors}.

\item
Mismatching of features which describe certain representational
properties for categories, as wrong head-argument relations, word order
and substitution of certain categories. These mismatchings produce {\em
structural errors}.

\end{enumerate}

Table~\ref{tab:1} shows the percentages of different types of errors
found in the corpus. Punctuation errors must be considered as structural
violations, while for style weaknesses, it depends on its subtype.
Errors at the lexical level are difficult to classify and most of them
must regarded as spelling rather than grammar errors. The number of
errors identified in this corpus is 543. These statistics could be
regarded as a representative average of the frequency of
errors/mistakes occurring in Spanish texts.

\begin{table}[htbp]
\caption{Statistics of errors}
\label{tab:1}

\tiny
\begin{minipage}{7.6cm}
\begin{tabular}{||p{2.4cm}|p{2.4cm}|p{1.3cm}||} \hline
\multicolumn{2}{||l|}{Type of error}         &    Percentage \\ \hline \hline
\multicolumn{2}{||l|}{Non-structural violations as described above}
                                             &
					\multicolumn{1}{r||}{18.5} \\ \hline
\multicolumn{2}{||l|}{Structural violations as described above}
                                             &
					\multicolumn{1}{r||}{9.7} \\ \hline
Punctuation    & Omission                    &
					\multicolumn{1}{r||}{32.2} \\
               & Addition                    & 
					\multicolumn{1}{r||}{4.8} \\ \hline
Errors at the  & At the character level      &
					\multicolumn{1}{r||}{6.3} \\
lexical level\footnote{Errors at the lexical level include typing
errors, word segmentation ({\em si no} vs {\em sino}), and cognitive
errors ({\em onceavo} (partitive) vs {\em und\'ecimo} (ordinal).}
               & Stress                      & \multicolumn{1}{r||}{8.0} \\ \hline
Style          & Structural                  &
					\multicolumn{1}{r||}{3.5} \\
weaknesses     & Lexical                     &
					\multicolumn{1}{r||}{12.0} \\ \hline
\multicolumn{2}{||l|}{Others}                &
					\multicolumn{1}{r||}{5.0} \\ \hline
\end{tabular}
\end{minipage}
\end{table}

A presupposition adopted in the project led to the idea that violations
at the feature level can be captured by means of the relaxation of the
possibly violated features while violations at the level of
configuration may not be relaxed without raising unpredictable parsing
results, thus being candidates for the implementation of explicit rules
encoding such incorrect structures.

Under this view, a comprehensive grammar checker must make use of both
strategies, called in the literature {\bf feature} or {\bf constraint
relaxation} and {\bf error anticipation}, respectively.

However, given the relevance of features in the encoding of linguistic
information in TFSs, some structural errors can be reanalyzed as
agreement errors in a wide sense (as feature mismatching violations
rather than structural ones). This allows the implementation of a
uniform approach to grammar correction, thus avoiding explicit rules
for ill-formed input. This paper describes such implementation for both
non-structural and structural violations.

\section{Error detection, diagnosis and correction techniques}

The overall strategy for detection, diagnosis and correction of
grammar and style errors within GramCheck relies on three axes:

\begin{itemize}
\item
For detection, a combined {\em feature relaxation} and {\em error
anticipation} approach is adopted. In order to implement the former,
extensive use of external CSs is performed in the analysis grammar,
whereas for the latter, explicit rules, adequately defined either in the
core grammar or in satellite subgrammars, are implemented\footnote{GramCheck
checks texts belonging to the standard language and to the
administrative sublanguage. The analysis module has been conceived as
composed by a core grammar and two satellite subgrammars ---for
overlapping cases--- that are mutually exclusive. Thus, the activation
of one subgrammar implies the deactivation of the other, and in both
cases they are added to the core grammar depending of the type of
(sub)language selected by the user.}.

\item
Diagnosis is performed by the CSs themselves with the aid of a {\em
heuristic technique} for those errors where tests should be performed
on several elements and a {\em pattern-related technique} which
provides a means to extend feature values with a gradation of correct
and posible but incorrect information. The typical case for the former
is agreement, thus for signs involving this type of information, both
an initial heuristic value is assigned and arithmetical operations are
performed on (in)equality tests. As for the latter, head-argument
relations where bound prepositions are involved are treated this way.
For all grammar errors there is no notion of weak vs. strong diagnosis,
being all considered strong errors needing automatic correction.

\item
Correction is performed by means of tree transduction of Linguistic
Structures (LSs) containing errors to a `language' (actually a `language
use') defined as {\em correct Spanish}. These synthesized structures are
displayed to the user. The overall design is then similar to a
transfer-based MT system, where the usual cycle is
analysis-transfer-synthesis, being the main differences the addition of
the above-mentioned grammar correction devices and the fact that not
all, but only incorrect sentences, will be pushed through the complete
cycle.

\end{itemize}

CSs allow the relaxation of certain features in the grammar rules whose
unification will be decided upon, in a non-trivial way, within these
CSs. Thus, rules do not perform feature value checking, so CSs play a
crucial role performing extended variable unification and taking
appropriate decisions.  Depending on the error type, CSs carry out
different operations on features, scores and lists.  These operations
concern basically the detection and the evaluation of the error,
providing a diagnostic on the error and correct value(s) for features
involved.  The use of CSs favours a one step diagnosis procedure where
decisions are only taken once all the relevant information is gathered.

\subsection{A heuristic technique to diagnose non-structural
errors}

CSs are used in GramCheck roughly in the way presented in
\cite{Crouch94} and \cite{Ruessink94}, developers of CSs for ALEP.
Nevertheless, while these reports describe constraint solving as an
extension to the expressive power of grammar rules, the novelty of the
approach presented here is the use of CSs to allow feature relaxation
in rules and boolean, relational and arithmetical operations on
relevant features. \\

Agreement errors pose a problem for a grammar checker which parses
natural language, because the detection of the error and the diagnosis
procedure have to be performed automatically. In inflectional
languages, like Spanish, this issue is essential given that in certain
contexts it is not possible to give a single correction when performing
analysis only at sentence level (i.e. without anaphoric relations). For
these cases, the system should be provided with a heuristics for the
correction in order to detect and diagnose the place(s) where the
error(s) has(have) been produced and to take a decision about the unit(s)
to be corrected. For GramCheck, this heuristics relies on a
parametrization of two assumptions: 

\begin{enumerate}
\item
the constituent which holds the feature values that in a given error
situation control the rest of the feature values in the other
constituents,

\item
the evaluation of the number of constituents which share and do not
share the same values.

\end{enumerate}

Our diagnosis procedure assumes that the gender and number features in
the head of a phrase control those in the dependent constituent(s),
although, as it will be proved later, this is not necessarily true. In
order to do this diagnosis procedure, the CS will contrast those
features and leave some clues of this evaluation in phrasal projections
in order for these to be available for further operations should they
were necessary. These clues are shaped as scores in the approach
adopted for agreement errors, and, in this sense, our heuristics is
closed to the metric operations performed by other grammar checkers
based on NLP techniques \cite{Veronis88}, \cite{Bolioli92},
\cite{Vosse92}, \cite{Genthial94}.

The core of this heuristics is that depending on a set of linguistic
principles based on lexico-morphological properties, the values for
gender and number in certain lexical units will be promoted over the
values in other units, thus, assigning them a higher score.

There are several conditions which have to be taken into account in
order to perform the diagnosis procedure. For instance, nouns with
inherent gender should control the gender of the rest of the elements
in a given NP. However, if the noun does not have inherent gender
---it's a noun that shows sex inflection--- then the gender value
should be controlled by those elements that, sharing the same value,
are majority. Hence, a sequence like {\em el}\_masc {\em casa}\_fem
(the house) must be corrected into {\em la}\_fem {\em casa}\_fem
because this noun has inherent feminine gender in Spanish. On the other
hand, an NP like {\em la}\_fem {\em chic-o}\_masc {\em guap-a}\_fem
(lit.  'the boy beautiful') should be corrected as {\em la}\_fem {\em
chic-a}\_fem {\em guap-a}\_fem ('the girl beautiful'), thus changing
the gender value of the head noun in the direction suggested by the
other dependent elements. This means that although the system could
take the gender value of the head as the value which commands the whole
phrase, the number of elements that share the same feature values, if
in contrast to those of the head and if the head takes its agreement
properties from morphology ---ie. are susceptible of keystroke
errors---, can influence the final decision.  Finally, for cases where
equal scores are obtained, as it happens with a non-inherent
masculine noun and a feminine determiner, both possible corrections
should be performed, since there is not enough information so as to
decide the correct value (unless this can be obtained from other
agreeing elements in the sentence ---for instance an attribute to this
NP).

Basically, the final operation to be performed with the scores is to
determine that the higher the score of an element the severer its
substitution. Thus, scores are clues for the correction of those
elements having the lowest scores. 

The initialization steps in order to perform the heuristic technique
are related to the assignment of values and scores to lexical
projections depending on its inherentness. The values for gender and
number of the head of the projection serve as a parameter for the
computation of values and scores for the possible modifier which could
appear closed to it. Note that agreement in Spanish is based on a
binary value system. Thus, the computation of values for the modifier
of a given head simply relies on the instantiation of opposite values
to those of the head. In the case of underspecification of the head for
gender, for instance, the presupposition is that this value is the same
as the one of the modifier, if this is not underspecified.  Otherwise,
both elements remain underspecified. Besides, the weight given to
controlling elements (50) ensures that there is no way for modifiers to
overpass this score. Note as well that the weight given to inherentless
values, as number (10), ensures that there are no promoted elements in
this calculation. The following schematic CS illustrates the
assignment of scores:

\tiny
\begin{verbatim}
and(=(Score_number_head,10),
  and(
    or(and(=(Inherentness_head,yes), =(Score_gender_head,50)),
       and(=(Inherentness_head,no), =(Score_gender_head,10))),
    =(Score_number_mod,0)))
\end{verbatim}

\normalsize
The following steps to be performed by CSs are related to the addition
of all those scores associated to a given value in the successive rules
building the nominal projection and the percolation of morphosyntactic
features:

\tiny
\begin{verbatim}
or(
   and(or(=(Gender_head_mother,Gender_mod),
          =(Gender_mod,masc_fem)),
      and(num_add(MGEN_SCORE_HEAD,
                  MGEN_SCORE_MOD,
                  MGEN_SCORE_MOTHER),
         and(=(Gender_mod_head,Gender_mod_mother),
            num_add(HGEN_SCORE_HEAD,
                    HGEN_SCORE_MOD,
                    HGEN_SCORE_MOTHER)))),
   and(=(Gender_mod,Gender_mod_head),
      and(num_add(HGEN_SCORE_HEAD,
                  MGEN_SCORE_MOD,
                  HGEN_SCORE_MOTHER),
         and(=(Gender_mod_head,Gender_mod_mother),
            num_add(MGEN_SCORE_HEAD,
                    HGEN_SCORE_MOD,
                    MGEN_SCORE_MOTHER)))))
\end{verbatim}

\normalsize
The final evaluation performed by CSs is done when categories
showing agreement overpass their maximal projection, only if no other
inter-syntagmatic agreement must be taken into account (as it is the
case with subject-attribute agreement, for instance).  Postponing in
this way the final evaluation ensures that the CS will take into
account all the previous parameters to give an appropriate diagnosis
about the complete XP containing the agreement violation. This evaluation
is based on the comparison of scores by means of the `greater than'
predicate in order to determine (a) the correct value for the feature(s)
checked corresponding to the highest score(s) ({\tt Right\_Gender,
Right\_Number} in the example below), to be used by the transfer module,
and (b) the error diagnosis ({\tt gender}, {\tt number} and {\tt
gender\_number} below), to be used by the error handling module that
will display appropriate error information to the user:

\newpage
\tiny
\begin{verbatim}
and(
or(
    or(and(num_gt(HGEN_SCORE_NOUN,MGEN_SCORE_NOUN),
               =(Gender_Noun,Right_Gender)),
       and(num_gt(MGEN_SCORE_NOUN,HGEN_SCORE_NOUN),
               =(Gender_Mod,Right_Gender))),
    =(HGEN_SCORE_NOUN,MGEN_SCORE_NOUN)),
or(
    or(and(num_gt(HNUM_SCORE_NOUN,MNUM_SCORE_NOUN),
              =(Number_Noun,Right_Number)),
       and(num_gt(MNUM_SCORE_NOUN,HNUM_SCORE_NOUN),
              =(Number_Mod,Right_Number))),
    =(HNUM_SCORE_NOUN,MNUM_SCORE_NOUN))),
\end{verbatim}

\normalsize
If all elements agree, scores for one of the arguments will always be
{\tt 0}, while if this argument has a value different than {\tt 0},
this information is considered as an evidence that an error has
occurred, the subsequent comparison determining the value for the
winning score:

\tiny
\begin{verbatim}
or(
   and(=(MNUM_SCORE_MOTHER,0), 
       or(=(MGEN_SCORE_MOTHER,0),
          and(num_gt(MGEN_SCORE_MOTHER,0),
              =(ERTYPE,gender)))),
   and(num_gt(MNUM_SCORE_MOTHER,0),
       or(and(=(MGEN_SCORE_MOTHER,0),
              =(ERTYPE,number)),
          and(num_gt(MGEN_SCORE_MOTHER,0),
              =(ERTYPE,gender_number)))))
\end{verbatim}

\normalsize
\subsection{A pattern-related technique to perform structural error
detection/diag\-nosis}

Turning back to the general definitions on error types given at the
beginning of this document, structural violations can be seen as
special cases of feature mismatching produced by addition, substitution
and omission of elements which result in a wrong dependency relation:

\vspace{2mm}

\noindent
{\bf Wrong head-argument relations} \\

\noindent
(i) Substitution of a bound preposition by another one (PP $\mapsto$
PP) \\

{\em Los alumnos relacionan la tarea [*a/con] su conocimiento}. \\

\noindent
(ii) Omission of a bound preposition resulting in a change of the
subcategorized argument (PP $\mapsto$ NP/S) \\

{\em Se acord\'{o} [*/de] que ten\'{\i}a una reuni\'{o}n por la
ma\~{n}ana.} \\

\noindent
(iii) Addition of a preposition resulting in a change of the
subcategorized argument (NP/S $\mapsto$ PP) \\

{\em Las empresas demandan [*de] m\'etodos.} \\

In the HPSG-like grammar used, bound prepositions are considered NPs
attached to the {\tt subcat} list (ie. the subcategorization feature)
of a predicative unit. These NPs have the feature {\tt pform}
instantiated to the value of the preposition, if any. If the argument
does not have a bound preposition, the value for {\tt pform} is {\tt
none}.  Thus, the approach adopted within GramCheck is that these error
cases have a correct representation of the dependency structure where
the only offending information is stored as a feature in the governed
element.  \\

The linguistic principle behind the pattern-related technique is based
on the fact that native writers substitute a preposition by another one
when certain associations between patterns, showing either the same
lexico-semantic and/or syntactic properties, are performed.  Thus, this
kind of error is not so accidental as it could be imagined.

For instance, Spanish speakers/writers usually associate the argument
structure of the comparative adjective {\em inferior} (lower), which
subcategorizes the preposition {\em a} (to), with the Spanish
comparative syntactic pattern ({\em menos ... que}, less ... than)
whose second term is introduced by the conjunction {\em que}, producing
phrases such as *{\em inferior que} instead of {\em inferior a}. With
the verb {\em relacionar} (to relate), something similar occurs: this
verb subcategorizes for the preposition {\em con}; however, due to the
fact that there exists the prepositional multi-word units {\em en
relaci\'on {\bf a}} and {\em en relaci\'on {\bf con}}, speakers tend to
think that the same prepositional alternation can be performed with the
verb (*{\em relacionar a} vs. {\em relacionar con}).

Following this idea, configurational rules are regarded, for grammar
checking, as descriptions of patterns, each of them having associated a
wrong pattern linked to the correct pattern. Both patterns are in a
complementary distribution. This way, structural errors can be foreseen
and controlled, and the system is provided with a mechanism which
establishes the way rule constraints must be relaxed.

To cope with this error, a CS operating on lists checks whether the
preposition in the constituent attached to the predicative sign belongs
to the head of the list or to the tail. If the preposition is member of
the tail, the same actions shown for agreement errors are performed
---instantiation of the correct value and determination of the error
type.

\section{Error coverage}

The current version of the GramCheck demonstrator is able to deal with
the following types of errors:

\begin{itemize}

\item
Intra- and inter-syntagmatic agreement errors (gender and/or number in
active ---with both predicative and copulative verbs--- and passive
sentences).

\item
Direct objects: omission of the preposition {\em a} with an animate entity
and addition of such a preposition with a non-animate entity.

\item
Addition, omission and substitution of a bound preposition covering
what is called {\em deque\'{\i}smo} ---the addition of a false bound
preposition {\tt de} with clausal arguments--- and {\em que\'{\i}smo}
---the omission of the bound preposition {\tt de} with clausal
arguments.

\item Errors on {\em portmanteau} words (use {\em de el, a el} instead
of {\em del, al}).

\end{itemize}

Regarding style issues, three different types of weaknesses are
detected: structural weaknesses, lexical weaknesses and abusive use of
passive, gerunds and manner adverbs. While structural weaknesses are
detected in the phrase structure rules using CSs (noun + ``a'' +
infinitive), by means of an error anticipation strategy, lexical
weaknesses are detected at the lexical level, with no special mechanisms
other than simple CSs. Lexical errors currently detected are related
with the use of Latin words which it is better to avoid, foreign words
with Spanish derivation, cognitive errors, foreign words for which a
Spanish word is recommended and verbosity.

\section{Further developments}

Results obtained with the current demonstrator are very promising. The
performance of the system using CSs is similar to that shown without
them, hence its use in conjunction with the detection techniques
proposed, rather than a burden, may be seen as a means to add
robustness to NLP systems. In fact, CSs may provide more natural
solutions to grammar implementation issues, like PP-attachment control.

Several directions for further developments have already been defined.
These include the integration of these grammar checking techniques
into the final release of the LS-GRAM Spanish grammar, which will have a
more realistic coverage in terms both of linguistic phenomena and
lexicon. Besides, on this new version of the grammar, hybrid techniques
will be used, taking advantage of the preprocessing facilities included
in ALEP. In particular, while for errors like those presented in this
paper the approach adopted is linguistically motivated, for certain
punctuation errors (or simply in order to reduce lexical ambiguity)
other relatively simple means can be defined that include certain
extended pattern matching on regular expressions or the passing of
linguistic information gathered in a preprocessing phase to the
unification-based parser. It is also foreseen to include a treatment for
{\em cognitive spelling errors}, usually not dealt with by conventional
spelling checkers.

\small
\bibliographystyle{acl}

\end{document}